\documentclass{article}

    \usepackage[preprint]{neurips_2024}

\usepackage[utf8]{inputenc} % allow utf-8 input
\usepackage[T1]{fontenc}    % use 8-bit T1 fonts
\usepackage{hyperref}       % hyperlinks
\usepackage{url}            % simple URL typesetting
\usepackage{booktabs}       % professional-quality tables
\usepackage{amsfonts}       % blackboard math symbols
\usepackage{amsthm}         % theorem environments
\usepackage{mathrsfs}
\usepackage{nicefrac}       % compact symbols for 1/2, etc.
\usepackage{microtype}      % microtypography
\usepackage{xcolor}         % colors
\usepackage[accsupp]{axessibility}
\usepackage{times}
\usepackage{graphicx,xcolor}
\usepackage{float}
\usepackage{subcaption}
\usepackage{amsmath,amssymb,bm,mathtools}
\usepackage{algorithm,algpseudocode}
\usepackage{dcolumn,colortbl,adjustbox,multirow,setspace}
\usepackage{cleveref}
\usepackage{natbib}       % 必须用numbers参数
\graphicspath{{figures/}}

\makeatletter%
\newcolumntype{L}{D{.}{.}{2,2}}
\newcolumntype{B}[3]{>{\boldmath\DC@{#1}{#2}{#3}}c<{\DC@end}}

\makeatother%

\newcommand\yulu{{{\text{RCR-AF}}}}

\title{RCR-AF: Enhancing Model Generalization via Rademacher Complexity Reduction Activation Function}

\author{%
  Yunrui Yu \\
  Tsinghua University \\
  \texttt{yuyunrui@mail.tsinghua.edu.cn} \\
  \And
  Kafeng Wang \\
  Tsinghua University \\
  \texttt{wangkafeng@mail.tsinghua.edu.cn} \\
  \And
  Hang Su \\
  Associate Professor \\
  Computer Science, Tsinghua University \\
  \texttt{hangsu@mail.tsinghua.edu.cn} \\
  Jun Zhu \\
  Professor \\
  Computer Science, Tsinghua University \\
  \texttt{junzhu@mail.tsinghua.edu.cn} \\
}

\begin{document}

\maketitle

\begin{abstract}
Despite their widespread success, deep neural networks remain critically vulnerable to adversarial attacks, posing significant risks in safety-sensitive applications. This paper investigates activation functions as a crucial yet underexplored component for enhancing model robustness. We propose a Rademacher Complexity Reduction Activation Function (\(\yulu\)), a novel activation function designed to improve both generalization and adversarial resilience. \(\yulu\) uniquely combines the advantages of GELU (including smoothness, gradient stability, and negative information retention) with ReLU's desirable monotonicity, while simultaneously controlling both model sparsity and capacity 
through built-in clipping mechanisms governed by two hyperparameters, 
$\alpha$ and $\gamma$. Our theoretical analysis, grounded in Rademacher complexity, demonstrates that these parameters directly modulate the model's Rademacher complexity, offering a principled approach to enhance robustness. Comprehensive empirical evaluations show that \(\yulu\) consistently outperforms widely-used alternatives (ReLU, GELU, and Swish) in both clean accuracy under standard training and in adversarial robustness within adversarial training paradigms.
\end{abstract}
\section{Introduction}
\label{sec:intro}

Deep learning has revolutionized artificial intelligence, achieving groundbreaking results across computer vision~\cite{he2016deep,krizhevsky2017imagenet}, natural language processing~\cite{vaswani2017attention}, and speech recognition~\cite{amodei2016deep}. However, the discovery of adversarial vulnerabilities~\cite{szegedy2013intriguing} has revealed a critical weakness: carefully crafted imperceptible perturbations can deceive even state-of-the-art models. This security flaw poses substantial risks as deep learning systems become increasingly deployed in safety-critical applications. The emergence of large language models like ChatGPT~\cite{achiam2023gpt} further amplifies these concerns, as their widespread adoption could magnify the societal impact of robustness failures.

Significant efforts have been made to develop defense mechanisms~\cite{goodfellow15,moosavi2016deepfool,carlini17,madry18,dong18momentum,xiao2019advgan,croce20aa} against adversarial attacks, alongside increasingly sophisticated evaluation methodologies~\cite{madry18,shafahi19free,alayrac19,zhang19trades,pang20hypersphere,wang20misclass,wu20wp,wu20width,yu2021lafeat,gao2022mora,yu2023lafit}. However, the arms race between evolving attack strategies and defense mechanisms underscores the critical need for \textit{theoretically grounded} approaches to robustness enhancement.

Defense strategies fall into three main categories: (1) \textit{adversarial training-based approaches}~\cite{madry18, zhang19trades}, which optimize models using clean and adversarial examples; (2) \textit{input transformation methods}~\cite{xiao2019advgan, dong18momentum}, which purify inputs to remove perturbations; and (3) \textit{model enhancement techniques}~\cite{wu20width,pang2020boosting}, which focus on architectural improvements and feature space regularization to inherently strengthen model robustness. Among these, adversarial training stands out as the most empirically validated, enhancing robustness by exposing models to adversarial perturbations during training~\cite{madry18}.

Recent studies \cite{rebuffi2021fixing} highlight the crucial role of activation functions in adversarial robustness. GELU consistently outperforms ReLU in adversarial training due to its continuous differentiability and retention of negative inputs, enabling stable gradient flow and improved robustness. However, its non-monotonic behavior reduces interpretability. In contrast, ReLU's hard zero-clipping promotes sparsity but discards negative information, potentially limiting robustness.

We propose the $\yulu$ activation function, which synergistically combines GELU's desirable properties (smoothness , gradient stability, and negative information retention) with ReLU's monotonicity. Notably, $\yulu$ incorporates built-in clipping with two hyperparameters ($\alpha$, $\gamma$) that simultaneously control feature sparsity and regulate model capacity via Rademacher complexity. This dual mechanism enables both practical output bounding and theoretically-grounded complexity control. To our knowledge, $\yulu$ represents the first activation function that systematically enhances adversarial robustness through explicit complexity regularization.

Experiments demonstrate $\yulu$'s advantages over standard activation functions (GELU, ReLU, and Swish): under standard training, it achieves higher clean accuracy, while in adversarial training settings, it yields improved robustness against AutoAttack. These consistent improvements across different training paradigms suggest that $\yulu$ can simultaneously enhance both generalization and adversarial robustness.

\begin{enumerate}
    \item The $\yulu$ activation function synergistically combines GELU's desirable properties (smoothness, gradient stability, and negative information retention) with ReLU's monotonicity.
    
    \item $\yulu$ incorporates built-in clipping with two hyperparameters ($\alpha$, $\gamma$) that simultaneously control feature sparsity and regulate model capacity via Rademacher complexity.
    
    \item Theoretical framework proving $\yulu$'s hyperparameters directly modulate model's Rademacher complexity.
    
    \item Comprehensive experiments showing $\yulu$ outperforms baseline activations in both standard and adversarial training scenarios.
\end{enumerate}
\section{Preliminaries \& Related Work}
\label{sec:prelim}

\subsection{Taxonomy of Defense Strategies}

Modern adversarial defense strategies fall into three main categories: (1) Adversarial training-based approaches, the most empirically validated paradigm, which builds on PGD-AT \cite{madry2017towards} with subsequent improvements like TRADES' theoretical loss decomposition \cite{zhang2019theoretically}, MART's margin maximization \cite{wang2019improving}, and enhanced regularization \cite{gowal2020uncovering}; (2) Input transformation methods that modify inputs pre-processing, including GAN-based purification \cite{xiao2019advgan} and stochastic randomization \cite{xie2017mitigating}, offering computational efficiency but weaker robustness guarantees; and (3) Model enhancement techniques like specialized activations \cite{hendrycks2016gaussian}, robust architectures \cite{wu2020adversarial}, and feature regularization \cite{pang2020boosting}, where our work contributes through novel activation function design while maintaining adversarial training compatibility.

\subsection{Adversarial Training Framework}
Among these, adversarial training stands out as the most empirically validated, enhancing robustness by exposing models to adversarial perturbations during training. Given a classifier $f_{\boldsymbol{\theta}} : \mathbb{R}^{C \times H \times W} \rightarrow \mathbb{R}^K$ with parameters $\boldsymbol{\theta}$, the adversarial training objective can be formulated as a min-max optimization problem:

\begin{equation}
\min_{\boldsymbol{\theta}} \mathbb{E}_{(\mathbf{x},y)\sim\mathcal{D}} \left[ \max_{\|\mathbf{\delta}\|_p \leq \epsilon} L(f_{\boldsymbol{\theta}}(\mathbf{x} + \mathbf{\delta}), y) \right]
\end{equation}

where $\mathcal{D}$ represents the data distribution and $L$ denotes the classification loss. The inner maximization generates adversarial perturbations within an $\ell_p$-norm ball of radius $\epsilon$, typically implemented through Projected Gradient Descent (PGD) \cite{madry2017towards}:

\begin{equation}
\mathbf{\hat{x}}_{i+1} = \text{Proj}_{\mathcal{B}_\epsilon(\mathbf{x})} \left( \mathbf{\hat{x}}_{i} + \alpha_i \cdot \text{sign} \left( \nabla_{\mathbf{\hat{x}}_{i}} L(f_{\boldsymbol{\theta}}(\mathbf{\hat{x}}_{i}), y) \right) \right),
\label{eq:pgd}
\end{equation}
where $\mathbf{\hat{x}}_0 = \text{Proj}_{\mathcal{B}_\epsilon(\mathbf{x})} (\mathbf{x} + \mathbf{u})$ is the initial adversarial example, with $\mathbf{u} \sim \text{Uniform}[-\epsilon, \epsilon]$ being a random perturbation.
 $\alpha_i$ is the step size at iteration $i$.
 $\text{Proj}_{\mathcal{B}_\epsilon(\mathbf{x})}$ denotes the projection operator that ensures the adversarial example remains within the $\epsilon$-ball centered at $\mathbf{x}$ under the $\ell_p$-norm (typically $\ell_\infty$ or $\ell_2$).
 $\nabla_{\mathbf{\hat{x}}_{i}} L$ is the gradient of the loss function $L$ with respect to the adversarial example $\mathbf{\hat{x}}_{i}$.

\subsection{Activation Functions in Robust Learning}
Activation functions play a pivotal role in determining neural network behavior during adversarial training. Modern activation functions can be categorized into two distinct classes based on their gradient properties: (1) piecewise-linear functions with gradient discontinuities (e.g., ReLU), and (2) smooth functions with continuous gradients (e.g., GELU and Swish).

The ReLU activation ($\sigma(x) = \max(0,x)$)~\cite{krizhevsky2017imagenet} represents the first category, introducing sparsity and computational efficiency through its simple thresholding operation. However, its non-differentiability at zero and complete suppression of negative activations can lead to training instability and dead neurons. These characteristics become particularly problematic in adversarial training scenarios, where gradient-based attacks exploit the discontinuous decision boundaries created by ReLU's hard thresholding.

In contrast, both GELU ($\sigma(x) = x\Phi(x)$ where $\Phi$ is the standard normal CDF)~\cite{hendrycks2016gaussian} and Swish ($\sigma(x) = x/(1+e^{-\beta x})$ with $\beta = 1$)~\cite{ramachandran2017searching} belong to the second category, providing continuous gradients across their entire domains. This shared smoothness property enables more stable gradient flow during both standard and adversarial training, while their retention of negative activations prevents complete neuron deactivation. Recent work by \cite{rebuffi2021fixing} demonstrates that these properties collectively contribute to improved adversarial robustness compared to ReLU.

\section{Methods}\label{sec:method}

\begin{figure}[htbp]
    \centering
    \includegraphics[width=0.8\linewidth]{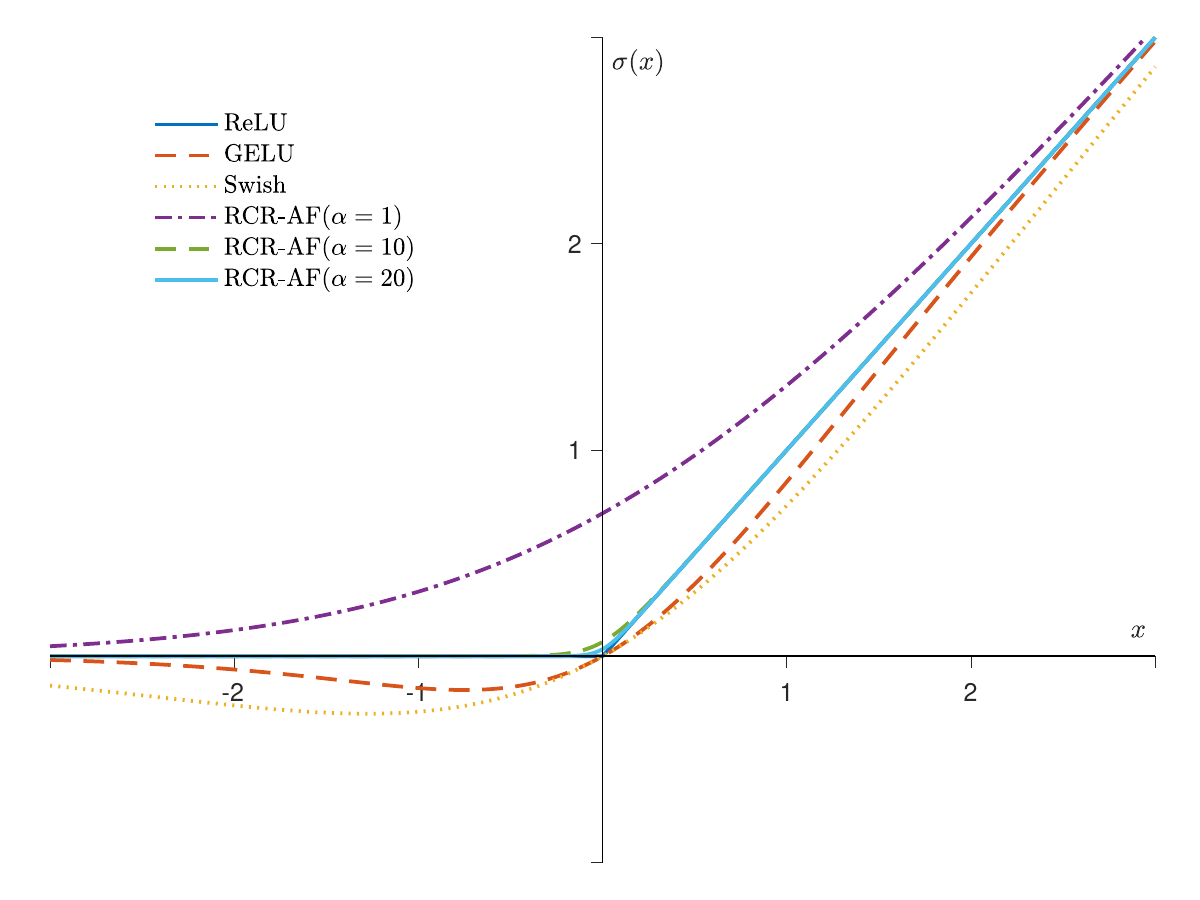}
    \caption{
    \textbf{Comparison of nonlinear activation functions:} ReLU ($\sigma(x)\!=\!\max(0,x)$),
    GELU ($\sigma(x)\!\approx\!x/(1\!+\!e^{-1.702x})$), Swish ($\sigma(x) = x/(1+e^{-\beta x})$ with $\beta = 1$), and $\yulu$ ($\sigma(x)\!=\!\alpha^{-1}\ln(1\!+\!e^{-\alpha x})\!+\!x$)
    with $\alpha\!\in\!\{1,10,20\}$. $\yulu$ preserves GELU's smoothness and Swish's gradient continuity while maintaining ReLU's strict monotonicity - a property neither GELU nor Swish possess. 
    As $\alpha$ increases to 10 (green dash-dot) and 20 (yellow dashed), $\yulu$ asymptotically approaches ReLU's piecewise linearity while retaining differentiability. 
    The $\yulu$ family thus provides a unified framework combining: (1) GELU's and Swish's  smoothness, gradient stability, and negative information retention, 
    (2) ReLU's guaranteed monotonicity, and (3) explicit control over nonlinearity through $\alpha$.
    }
    \label{fig:yulu_activation}
\end{figure}

We propose a novel activation function, termed $\text{\(\yulu\)}(x; \alpha;\gamma)$, defined as:

\begin{equation}
\text{\(\yulu\)}(x; \alpha;\gamma) = 
\frac{1}{\alpha}\ln\left(1+e^{-\alpha x}\right)+x  \quad with \quad 
\text{clip}(  x, [-\gamma/\alpha, \gamma/\alpha])
\end{equation}

where $\alpha$ and $\gamma$ are positive parameters. Without considering the clipping operation, \cref{fig:yulu_activation} illustrates $\yulu$'s behavior compared to ReLU, GELU, and Swish for $\alpha$ values of 1, 10, and 20. Notably, $\yulu$ preserves the smoothness, gradient stability, and negative information retention of GELU, while maintaining the monotonicity of ReLU. As $\alpha$ increases, $\yulu$ progressively converges toward ReLU's characteristics. When $\alpha=20$, $\yulu$ becomes nearly indistinguishable from ReLU in the plotted range.

\subsection{Model Sparsity Controlled by Built-in Clipping Hyperparameters}
\label{sec:sparse_1}

The derivative of \(\sigma_{\text{\(\yulu\)}}(x)\) is:
\begin{equation}
\sigma'_{\text{\(\yulu\)}}(x) = \frac{d}{dx} \left[ \frac{1}{\alpha} \ln(1 + e^{-\alpha x}) + x \right] = \frac{1}{1 + e^{-\alpha x}}.
\label{eq:derivative}
\end{equation}
 
For large \(\alpha\), the term \(e^{-\alpha x}\) may overflow to infinity (inf) when:
\(e^{-\alpha x} > M,\)
where \(M\) is the maximum representable value (\(\approx 10^{4.85}\) for 16-bit, \(10^{38.23}\) for 32-bit, \(10^{308.25}\) for 64-bit). Overflow occurs when:
\(
x < -\frac{\lambda}{\alpha}, \quad \lambda = \ln M \approx 11.0899, 88.7228, 709.7827,
\)
for 16-bit, 32-bit, and 64-bit precision, respectively. In this regime:
\(
\frac{d}{dx} \sigma(x) = \frac{1}{1 + \text{inf}} = 0.
\)
Zero gradients prevent parameter updates, causing neurons to “die” and promoting sparsity in the activations, similar to ReLU’s dying neuron problem. 
The probability of this occurring for pre-activations \(z_{i,j} \sim \mathcal{N}(0, \sigma_j^2)\) is:
\begin{equation}
\mathbb{P}\left( z_{i,j} < -\frac{\lambda}{\alpha} \right) = \Phi\left( -\frac{\lambda}{\alpha \sigma_j} \right)
\label{eq:probability_of_die}
\end{equation}
where \(\Phi\) is the standard normal CDF.  The sparsity probability increases with larger values of $\alpha$. To achieve active control over both the model's sparsity and the maximum output value of individual neural network layers through the $\yulu$ activation function, we propose a modified implementation that incorporates a clipping operation $\text{clip}(x, [-\gamma/\alpha, \gamma/\alpha])$, where $0 < \gamma < \lambda$ represents a tunable parameter. This approach is motivated by the observation that the clipping operation naturally produces zero gradients for clipped values. After clipping, the probability of model sparsity occurring for pre-activations $z_{i,j} \sim \mathcal{N}(0, \sigma_j^2)$ is:

\begin{equation}
\mathbb{P}\left( |z_{i,j}| > \frac{\gamma}{\alpha} \right) = 2\Phi\left( -\frac{\gamma}{\alpha \sigma_j} \right)
\end{equation}

Meanwhile, the maximum output value of a single neural network layer through the $\yulu$ activation function becomes:

\begin{equation}
M_{\text{clip}} = \frac{1}{\alpha} \ln(1 + e^{-\gamma}) + \frac{\gamma}{\alpha}
\end{equation}
 
When $\gamma$ is fixed, $M_{\text{clip}}$ decreases monotonically with increasing $\alpha$. Simultaneously, when $\alpha$ is fixed, reducing $\gamma$ directly decreases $\gamma/\alpha$, thereby imposing stronger constraints on the maximum output values of each neural network layer.

\subsection{Rademacher Complexity Control via tunable parameters}
\label{sec:rademacher_control}

For any given model, reducing its Rademacher complexity effectively enhances generalization capability. Since model robustness constitutes a crucial aspect of generalization, our theoretical analysis primarily focuses on the Rademacher complexity perspective. We demonstrate that when using $\yulu$ as the activation function, one can directly control the model's Rademacher complexity through adjustment of the $\alpha$ and $\gamma$ parameters. This theoretical foundation explains how our novel activation function enhances model robustness.

Consider an \(L\)-layer network:
\[
f(x) = W_L \sigma(W_{L-1} \sigma(\cdots \sigma(W_1 x) \cdots )),
\]
Suppose that \(W_i \in \mathbb{R}^{d_{i-1} \times d_{i}}\),  \( \forall j \quad \|x^{(j)}\|_2 \leq c\), \(z_i=W_i x_i\), and \(\sigma_{\text{\(\yulu\)}}(z_i) \equiv \sigma_{\text{\(\yulu\)}}\left(\text{clip}(  z_i, [-\gamma/\alpha, \gamma/\alpha])\right)\). For layer \(i\), the function class is:
\[
\mathcal{F}_i = \{ f_{\theta} : \|W_i\|_{op} \leq k_i, \|W_i^T\|_{2,1} \leq b_{i} \}.
\]

\subsubsection{Lipschitz Constant of the Activation Function under Bounded Inputs}

Given the bounded input property of our activation function, where $|z| \leq \frac{\gamma}{\alpha}$, we can precisely compute the Lipschitz constant of \(\yulu\). The derivative of \(\yulu\) is given by \ref{eq:derivative}.
To determine the Lipschitz constant $L$, we find the maximum absolute value of the derivative within the bounded domain:

\begin{equation}
L = \sup_{|z| \leq \gamma/\alpha} |\sigma'(z)|
\end{equation}

For $z \in [-\gamma/\alpha, \gamma/\alpha]$, The maximum occurs at $z=\gamma/\alpha$ with $\frac{1}{1 + e^{-\gamma}}$. Therefore, the Lipschitz constant is:

\begin{equation}
L = \frac{1}{1 + e^{-\gamma}}<1
\end{equation}

\subsubsection{Operator Norm Analysis with Clip Operation}
\label{sec:norm_analysis}

% Given a fully-connected neural network layer with weight matrix $W_i \in \mathbb{R}^{d_i \times d_{i-1}}$, we analyze the operator norm after the clip operation $\text{clip}(W_ix, [-\gamma/\alpha, \gamma/\alpha])$. Under the initial constraints $\|W_i\|_{op} \leq k_i$ and $\|W_i^T\|_{2,1} \leq b_i$, we derive the modified operator norm bound as follows.

The clip operation, being a 1-Lipschitz function, satisfies $\|\text{clip}(z_1) - \text{clip}(z_2)\|_2 \leq \|z_1 - z_2\|_2$, which implies:
\begin{equation}
\|\text{clip}(W_ix)\|_2 \leq \|W_ix\|_2 \leq k_i\|x\|_2
\end{equation}

However, the clip operation $\text{clip}(W_ix, [-\gamma/\alpha, \gamma/\alpha])$ introduces an additional bound:
\begin{equation}
\|\text{clip}(W_ix)\|_2 \leq \|\text{clip}(W_i)\|_{op}\|x\|_2 \leq  \sqrt{d_i}\cdot\frac{\gamma}{\alpha}
\end{equation}

We obtain the final operator norm bound:

\begin{equation}
\|\text{clip}(W_i)\|_{op} = \sup \|\text{clip}(W_i x)\|_2 \leq \min\left(k_i, \sqrt{d_i} \cdot \frac{\gamma}{\alpha \|x\|_2}\right)
\end{equation}

Given that $\sigma_{\text{\(\yulu\)}}$ is $\frac{1}{1+e^{-\gamma}}$-Lipschitz, we derive:
\[
\|\sigma(\text{clip}(W_i x)) - \sigma(\text{clip}(W_i x'))\|_2 \leq \frac{1}{1+e^{-\gamma}}\|\text{clip}(W_i x) - \text{clip}(W_i x')\|_2 
% \leq \frac{1}{1+e^{-\gamma}}\|\text{clip}(W_i)\|_{\text{op}} \|x - x'\|_2.
\]
Consequently, we obtain the operator norm  with clip operation:
\[
\|\sigma(\text{clip}(W_i ))\|_{\text{op}} \leq \frac{\min\left(k_i, \sqrt{d_i} \cdot \frac{\gamma}{\alpha \|x\|_2}\right)}{1+e^{-\gamma}},
\]

% \paragraph{Compatibility with Existing Rademacher Complexity Analysis}

To maintain compatibility with the Rademacher complexity derivation presented in \cite{ma2020mltheory} (Chapter 5, pp. 62--67), we define the operator norm for the $i$-th network layer as:

\begin{equation}
k_{\text{i,clip}} = \zeta_{\text{clip}} k_i,
\end{equation}

where $\zeta_{\text{clip}} = \frac{\min\left(1, \sqrt{d_i} \cdot \frac{\gamma}{\alpha \|x\|_2 k_i}\right)}{1+e^{-\lambda}} < 1$ the clipping factor that captures the reduction in operator norm. 

This formulation allows direct substitution of $k_{\text{i,clip}}$ for the original operator norm $k_i$ derived under the assumption of ReLU activation (with Lipschitz constant 1). Consequently, we can directly apply the existing theoretical framework from \cite{ma2020mltheory} to analyze the Rademacher complexity of our model while accounting for the modified operator norm induced by our activation function.

\subsubsection{Clipping Effect on Covering Number}

\paragraph{Covering Number Bound for Unclipped \(\yulu\) Activation}
Theorem 5.18 from \cite{ma2020mltheory} like below:

Let \( \mathcal{F} = \{ x \to Wx : W \in \mathbb{R}^{m \times d}, \|W^T\|_{2,1} \leq B \} \) and let \( C = \sqrt{\frac{1}{n} \sum_{i=1}^n \|x^{(i)}\|_2^2} \). Then,

\begin{equation}
  \log N(\epsilon, \mathcal{F}, \|\cdot\|_2) \leq \left[ \frac{c^2 B^2}{\epsilon^2} \right] \ln(2dm).  
\end{equation}

As established in our previous analysis, the \(\yulu\) activation function without clipping operations has a Lipschitz constant of 1. Consequently, applying
we obtain the following covering number bound for each network layer when using unclipped \(\yulu\) as the activation function can be bounded by:

\begin{equation}
\log N(\epsilon_i, \mathcal{F}_i, \|\cdot\|_2) \leq   \frac{c_{i-1}^2 b_i^2}{\epsilon_i^2} \ln(2d_{i-1}d_{i}),
\end{equation}

where $c_{i-1}^2$ denotes the norm of the input features , $b_i^2 $ is $ \|W_i^\top\|_{2,1}$ ,and $\epsilon_i^2$ is the covering radius.

\paragraph{Output covering Number Bound for clipped version}
Since \(\text{clip}(  z, [-\gamma/\alpha, \gamma/\alpha])\), the output space is \((0, M_{\text{clip}}]^{d_{i}}\). 

\noindent The $\epsilon$-covering number under $\ell_2$-norm is derived as follows:

    \begin{equation}
    N(\epsilon_i, (0, M_{\text{clip}}]^{d_i}, \|\cdot\|_2) \leq \left[ \frac{M_{\text{clip}}\sqrt{d_i}}{\epsilon_i} +1\right]^{d_i}
    \end{equation}

Logarithmic Form:
    \begin{equation}
    \log N(\epsilon_i, (0, M_{\text{clip}}]^{d_i}, \|\cdot\|_2) \leq d_i \ln \left( \frac{M_{\text{clip}}\sqrt{d_i}}{\epsilon_i} + 1 \right) 
    % \leq  d_i \log \left( \frac{2M_{\text{clip}}\sqrt{d_i}}{\epsilon_i} \right)
    \end{equation}

\paragraph{Combined Bound}
A tighter bound combines the weight space covering with the output constraints:
\[
g_\text{clip}\left( \epsilon_i, c_{i-1} \right) = \min \left( \frac{b_i^2 c_i^2}{\epsilon_i^2}\ln(2d_{i-1}d_{i}), \quad d_{i} \ln \left( \frac{M_{\text{clip}}\sqrt{d_i} }{\epsilon_i} +1\right) \right) .
\]

To maintain compatibility with the Rademacher complexity derivation presented in \cite{ma2020mltheory} (Chapter 5, pp. 62--67), we define the clipped covering number for the $i$-th layer as:

\begin{equation}
g_{\text{clip}}\left( \epsilon_i, c_{i-1} \right) =  \frac{b_{\text{i,clip}}^2 c_i^2}{\epsilon_i^2} \ln(2d_{i-1}d_{i}),
\end{equation}

where $b_{\text{i,clip}} = \eta_{\text{i, clip}} b_i$ with $\eta_{\text{i, clip}} $ being the clipping factor that captures the reduction in output space complexity.

\begin{equation}
    \frac{b_{\text{i, clip}}^2 c_i^2}{\epsilon_i^2} \ln(2d_{i-1}d_{i}) = \min \left( \frac{b_i^2 c_i^2}{\epsilon_i^2}\ln(2d_{i-1}d_{i}), \quad d_{i} \ln \left( \frac{M_{\text{clip}}\sqrt{d_i} }{\epsilon_i} +1\right) \right)
\end{equation}

\begin{align}
\eta_{\text{clip}}^2 &= \min \left(1,  \frac{d_i \ln \left( \frac{M_{\text{clip}}\sqrt{d_i}}{\epsilon_i} +1\right) \epsilon_i^2}{b_i^2 c_i^2 \ln(2d_{i-1}d_{i})} \nonumber \right) \leq 1
% = \frac{d_i \ln \left( \frac{3\left(\frac{1}{\alpha}\ln(1+e^{-\gamma}) + \frac{\gamma}{\alpha}\right)\sqrt{d_i}}{\epsilon_i} \right) \epsilon_i^2}{b_i^2 c_i^2}.
\end{align}

This formulation enables direct substitution of $b_{\text{i,clip}}$ for the original $b_i$ in \cite{ma2020mltheory}'s framework, preserving the theoretical derivation while accounting for output space constraints. The clipping factor $\eta_{\text{clip}}$ quantitatively measures the complexity reduction induced by bounded activations.

\paragraph{Rademacher Complexity Analysis with \(\yulu\) Activation}
After substituting $k_{\text{i,clip}}$ for $k_i$ and $b_{\text{i,clip}}$ for $b_i$ following the derivation in \cite{ma2020mltheory} (Chapter 5, pp. 62--67), we obtain the modified Rademacher complexity bound:

\begin{equation}
\begin{split}  
R_S(\mathcal{F}) & \leq \frac{c}{\sqrt{n}} \cdot {\left( \prod_{i=1}^r k_{i,\text{clip}} \right)} \cdot {\left( \sum_{i=1}^r \frac{b_{i,\text{clip}}^{2/3}}{k_{i, \text{clip}}^{2/3}} \right)^{3/2}} \\
& =\frac{c}{\sqrt{n}} \cdot {\left( \prod_{i=1}^r (k_{i}\zeta_{\text{clip}}) \right)} \cdot {\left( \sum_{i=1}^r \frac{(b_{i}\eta_{i,\text{clip}})^{2/3}}{(k_{i}\zeta_{\text{clip}})^{2/3}} \right)^{3/2}}\\
& <\frac{c}{\sqrt{n}} \cdot {\left( \prod_{i=1}^r k_{i} \right)} \cdot {\left( \sum_{i=1}^r \frac{b_{i}^{2/3}}{k_{i}^{2/3}} \right)^{3/2}}.
\label{eq:Rademacher_complexity}
\end{split}
\end{equation}

Equation~\ref{eq:Rademacher_complexity} establishes that the \(\yulu\) activation function enables explicit control of the model's Rademacher complexity through its hyperparameters $\alpha$ and $\gamma$. When the parameter $\gamma$ is held constant, the Rademacher complexity exhibits a monotonic decreasing relationship with respect to $\alpha$. This property provides a direct mechanism for enhancing the model's generalization capability, as reduced Rademacher complexity typically leads to improved generalization bounds. Furthermore, through careful joint optimization of $\lambda$ and $\alpha$, we can systematically improve the model's robustness in adversarial training scenarios.

\section{%
Experiments%
}\label{sec:results}
The theoretical analysis in section \ref{sec:method} demonstrates that for the proposed \(\yulu\) activation function, the model's Rademacher Complexity exhibits a decreasing trend as parameter $\alpha$ increases while keeping parameter $\gamma$ constant. This section systematically evaluates the effectiveness of the \(\yulu\) activation function in reducing Rademacher Complexity and consequently improving model generalization capability by adjusting $\alpha$ with a fixed $\gamma$ value (throughout all experiments, we set $\gamma \approx 66.7228$). The comprehensive impact of different $\alpha$ and $\gamma$ combinations on model performance will be investigated in our future work. Model generalization capability is assessed through two primary metrics: (1) Clean Accuracy achieved on test datasets under standard training paradigm, and (2) robustness performance demonstrated in adversarial training scenarios.

\subsection{Clean Accuracy Evaluation under Standard Training}
To evaluate the impact of the \(\yulu\) activation function on clean accuracy in standard training scenarios, we conducted comparative experiments against mainstream activation functions (ReLU, GELU, SWISH) under identical training conditions including strong data augmentation (Google's AutoAugment). All experiments were performed on the CIFAR-10 dataset using ResNet-18 as the backbone architecture. All models were trained with standardized configurations: batch size of 128, 200 training epochs, and AutoAugment enabled throughout. For the \(\yulu\) activation function, we systematically tested its key parameter $\alpha$ across values ranging from 5 to 100. To mitigate randomness, each experiment was repeated three times with random initialization, with the maximum value among the three trials used for comparison.
\begin{figure}[htbp]
    \centering
    \includegraphics[width=0.8\linewidth]{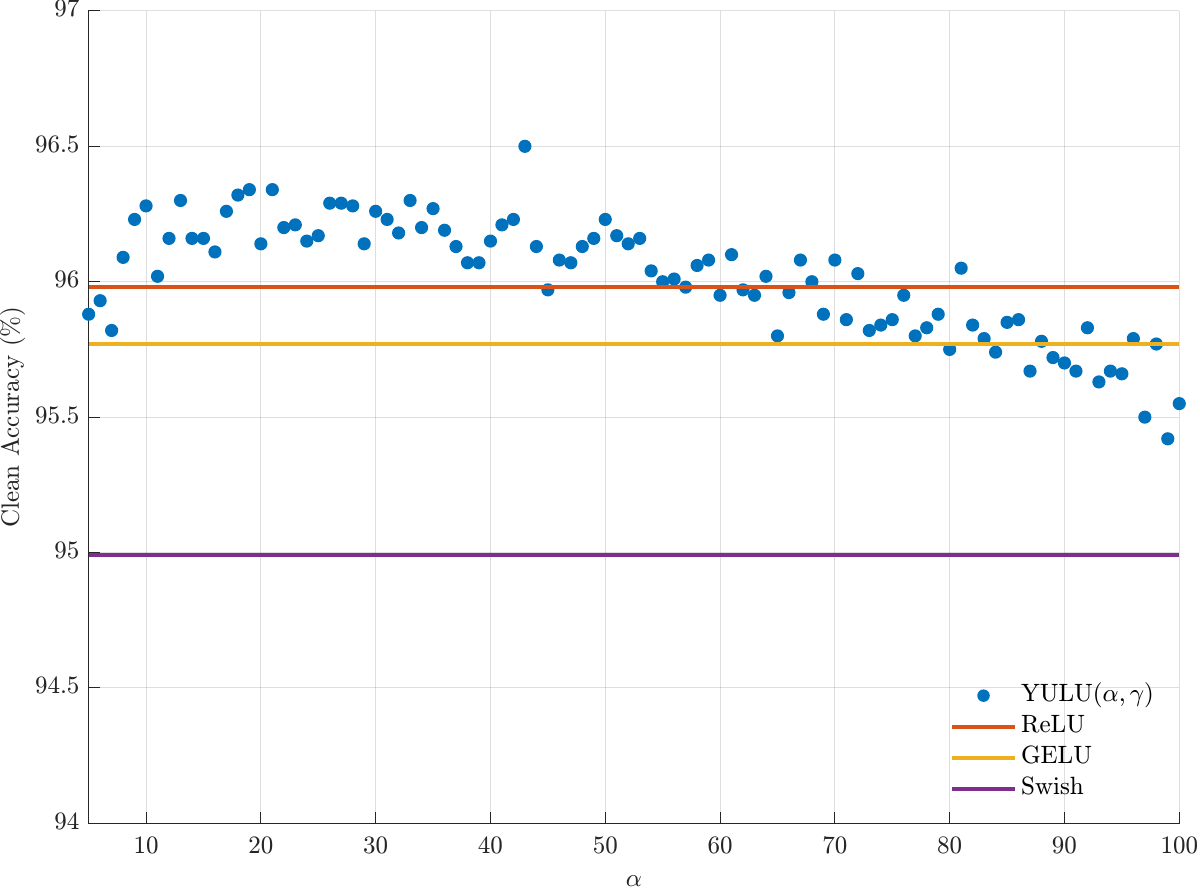}
    \caption{
    Clean accuracy comparison on CIFAR-10 (ResNet-18) with all values representing the best of three independent trials. The proposed \textbf{\(\yulu\)} ($\alpha \in [5,100]$) shows $\alpha$-dependent performance (blue markers), while baseline activations achieve: \textbf{ReLU} (95.98\%), \textbf{GELU} (95.77\%), and \textbf{Swish} (94.99\%).
    }
    \label{fig:clean_acc}
\end{figure}
As shown in \ref{fig:clean_acc}, when $\alpha$ increases from 5 to 20, the model's clean accuracy demonstrates a significant upward trend, indicating that increasing $\alpha$ within this range enhances generalization capability. Notably, with $\alpha$ set to 43, the ResNet-18 model equipped with \(\yulu\) activation achieves 96.50\% Clean Accuracy on the CIFAR-10 test set. Comparative analysis reveals \(\yulu\)'s superior performance: it outperforms ReLU (95.98\%) by 0.52\%, GELU (95.77\%) by 0.73\%, and SWISH (94.99\%) by 1.51\%. These results preliminarily confirm that the properly parameterized \(\yulu\) activation function can effectively improve classification accuracy within standard data augmentation pipelines.

However, when $\alpha$ exceeds 50, further increases lead to a noticeable decline in clean accuracy. This suggests that excessively large $\alpha$ values may over-constrain the model's representational capacity, potentially causing underfitting and ultimately degrading model performance.

\subsection{Robustness Evaluation under Adversarial Training}
To assess the impact of \(\yulu\) activation on model robustness, we followed the adversarial training framework proposed in \cite{rebuffi2021fixing}, training and evaluating ResNet-18 models with different activation functions using only the CIFAR-10 dataset without additional data. The training employed a batch size of 128 with weight averaging strategy over 400 epochs. The \(\yulu\) activation's parameter $\alpha$ was adjusted within the range of 5 to 50. Model robustness was evaluated using AutoAttack (AA).

\begin{figure}[htbp]
    \centering
    \includegraphics[width=0.8\linewidth]{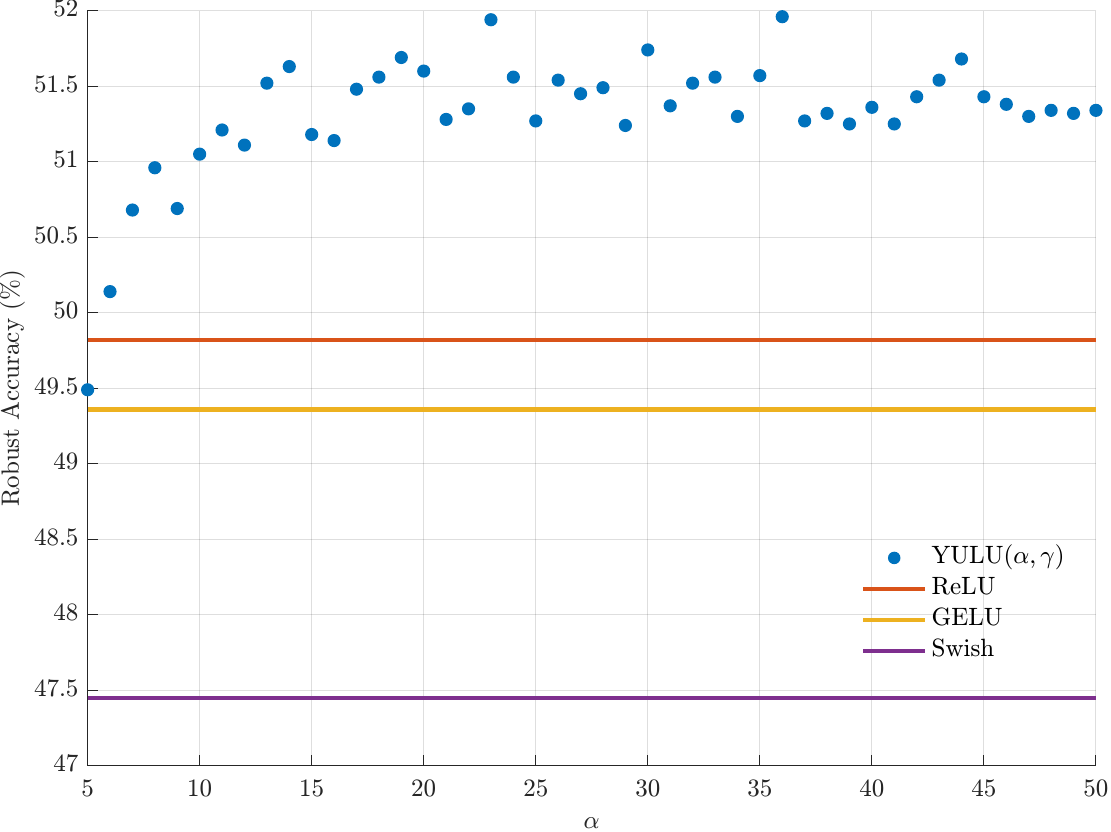}
    \caption{
    Robustness comparison of activation functions under PGD adversarial training on CIFAR-10 (ResNet-18). Baseline results for \textbf{ReLU} (adv. acc 49.82\%), \textbf{GELU} (49.36\%), and \textbf{Swish} (49.36\%) show the best of three independent trials, while \textbf{\(\yulu\)} ($\alpha\in[5,50]$) presents single-trial results (blue markers). \(\yulu\) achieves superior robust accuracy (peak 51.96\% at $\alpha=36$) compared to all baselines, demonstrating $\alpha$-dependent robustness characteristics.
    }
    \label{fig:adv_acc}
\end{figure}
\ref{fig:adv_acc} demonstrates that with $\alpha=36$, the \(\yulu\)-equipped model achieves 51.96\% robustness under AA evaluation. Comparative results show \(\yulu\)'s robustness advantages: it surpasses ReLU (49.82\%) by 2.14\%, GELU (49.36\%) by 2.60\%, and SWISH (47.45\%) by 4.51\%.

\subsection{Experimental Summary}
Our comprehensive experimental results demonstrate that with a fixed $\gamma$, selecting an appropriate $\alpha$ value for the \(\yulu\) activation function can effectively enhance model generalization capability. This improvement manifests in both increased Clean Accuracy under standard training and enhanced robustness under adversarial training. Across both scenarios, \(\yulu\) activation consistently outperforms widely-used alternatives including ReLU, GELU, and SWISH. These findings provide preliminary evidence for \(\yulu\) activation's potential in improving both generalization capability and robustness of deep learning models.
\section{Conclusion}\label{sec:conclusion}

This paper introduced \yulu, a novel activation function meticulously designed to enhance both the generalization capabilities and adversarial robustness of deep neural networks. Our theoretical analysis, grounded in Rademacher complexity, demonstrated that \yulu's unique hyperparameters, \(\alpha\) and \(\gamma\), provide a principled mechanism for directly influencing model capacity and, consequently, generalization bounds—a crucial aspect for developing robust AI. Comprehensive experimental evaluations substantiated these theoretical underpinnings. \(\yulu\) consistently outperformed widely adopted activation functions, including ReLU, GELU, and SWISH, delivering significant improvements in clean accuracy under standard training conditions and, critically, superior adversarial robustness within adversarial training paradigms. These findings not only highlight \yulu's potential as an immediate advancement for building more reliable and secure machine learning models but also contribute a methodological foundation for the future design of activation functions tailored to the increasing demands of robust AI in safety-critical applications and large-scale deployments.

\clearpage

{\small
\bibliography{references}
}
\clearpage
\section{Technical Appendices and Supplementary Material}

\subsection{Limitations of Hyperparameter Exploration}
\label{app:limitations}

While the proposed $\yulu$ activation function demonstrates significant performance advantages over existing alternatives in our experiments, this study has one key limitation regarding hyperparameter tuning. All experimental results were obtained with $\gamma$ held constant while varying $\alpha$. This partial exploration of the hyperparameter space leaves open questions about the joint optimization of both parameters.

This limitation deliberately narrows the scope of the current empirical validation to establish a clear baseline performance. Future work should systematically investigate reinforcement learning strategies for automated exploration of optimal $(\alpha, \gamma)$ combinations.

\subsection{Hardware Specifications for Experiments}
\label{app:hardware}

To ensure reproducibility of all experimental results, we provide complete hardware specifications for our computational environment. All experiments were conducted on a local workstation equipped with an NVIDIA GeForce RTX 2080 Ti GPU.

The approximate execution times for key experimental configurations were:
\begin{itemize}
    \item \textbf{Standard Training}: Approximately 2 hour per run
    \item \textbf{Adversarial Training}: Approximately 28 hours per run
\end{itemize}

These timings account for complete training cycles including model initialization, data loading, and checkpoint saving. All reported results were obtained using this local hardware configuration without additional cloud or cluster resources. Preliminary experiments (hyperparameter tuning and exploratory runs) were also performed on the same system but are excluded from final reported metrics.

\end{document}